\documentclass[final]{cvpr}

\usepackage{times}
\usepackage{epsfig}
\usepackage{graphicx}
\usepackage{amsmath}
\usepackage{amssymb}
\usepackage{algorithm}
\usepackage{bm}
\usepackage{multirow}

\usepackage[pagebackref=true,breaklinks=true,colorlinks,bookmarks=false]{hyperref}

\def\cvprPaperID{6} 
\def\confYear{CVPR 2021}

\begin{document}

\title{Embedded Knowledge Distillation in Depth-Level Dynamic Neural Network}

\author{
Qi Zhao\textsuperscript{1} \and Shuchang Lyu\textsuperscript{1} \and Zhiwei Zhang\textsuperscript{1}
\and Ting-Bing Xu\textsuperscript{1} \thanks{Corresponding author}
\and Guangliang Cheng\textsuperscript{2} \and \\
\textsuperscript{1} Beihang University \textsuperscript{2} SenseTime Research, Beijing \\
{\tt\small \{zhaoqi, lyushuchang, zhangzwq1998, tingbing\_xu\}@buaa.edu.cn}, \tt\small chengguangliang@sensetime.com
}



\maketitle

\begin{abstract}
  In real applications, different computation-resource devices need different-depth networks (e.g., ResNet-18/34/50) with high-accuracy. Usually, existing methods either design multiple networks and train them independently, or construct depth-level/width-level dynamic neural networks which is hard to prove the accuracy of each sub-net. In this article, we propose an elegant Depth-Level Dynamic Neural Network (DDNN) integrated different-depth sub-nets of similar architectures. To improve the generalization of sub-nets, we design the Embedded-Knowledge-Distillation (EKD) training mechanism for the DDNN to implement knowledge transfer from the teacher (full-net) to multiple students (sub-nets). Specifically, the Kullback-Leibler (KL) divergence is introduced to constrain the posterior class probability consistency between full-net and sub-nets, and self-attention distillation on the same resolution feature of different depth is addressed to drive more abundant feature representations of sub-nets. Thus, we can obtain multiple high-accuracy sub-nets simultaneously in a DDNN via the online knowledge distillation in each training iteration without extra computation cost. Extensive experiments on CIFAR-10/100, and ImageNet datasets demonstrate that sub-nets in DDNN with EKD training achieve better performance than individually training networks while preserving the original performance of full-nets.
\end{abstract}

\section{Introduction}
\begin{figure}[t]
  \centering
  \includegraphics[width=1.00\linewidth]{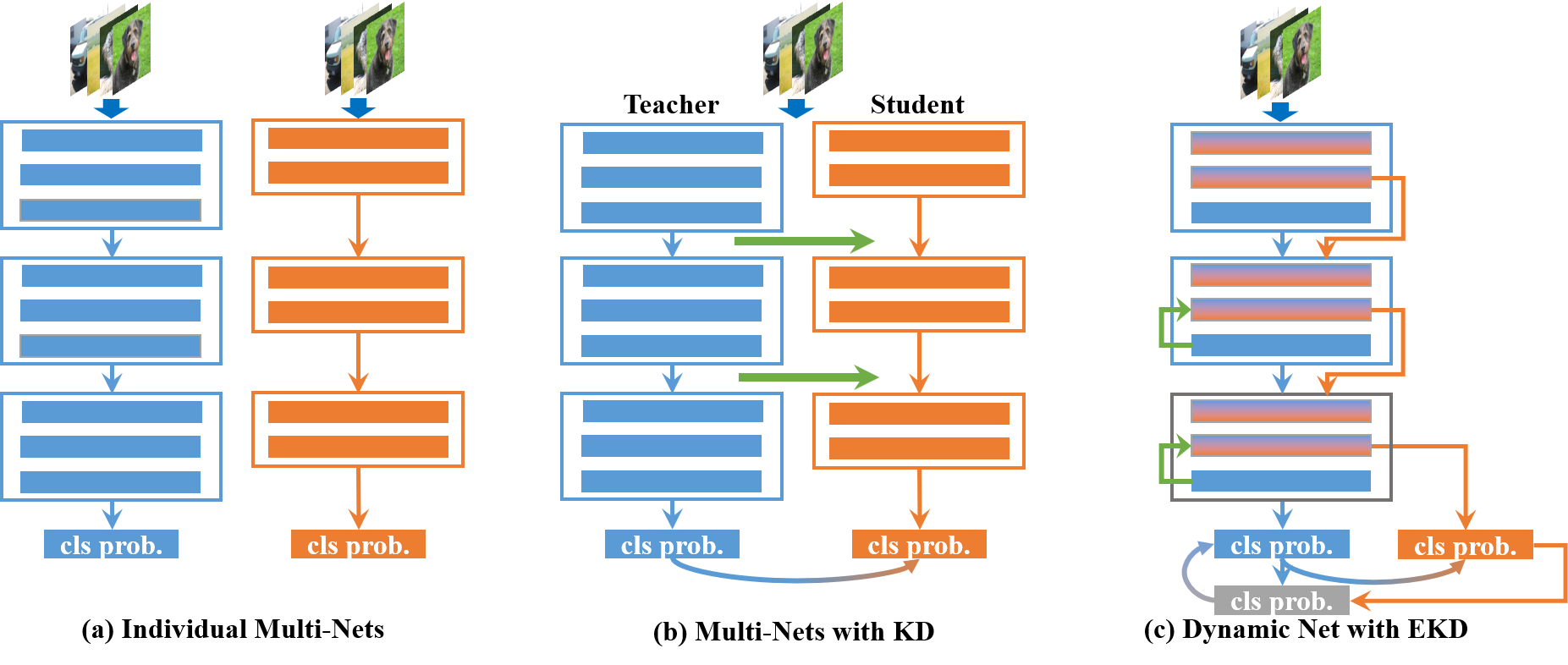}     \\
  \caption{Diagrams of three cases. (a) Individual multi-nets. (b) Teacher-to-student with KD. (c) DDNN with EKD.}\label{Fig1}
\end{figure}

Recent years have witnessed significant progress in various computer vision tasks\cite{ResNet,Faster_RCNN,FCN} using deep convolutional neural networks. To meet different resource-constrained devices, researchers usually need to design a series of different-depth networks such as VGG-13/16/19~\cite{VGG}, ResNet-18/34/50/101~\cite{ResNet}, and DensNet-121/169/201~\cite{DenseNet}. Generally, it requires to train different-depth networks individually in Fig.~\ref{Fig1}(a) and download/offload different models multiple times according to device-resource constraints in real applications, which increases the training and deploying cost dramatically. In fact, deeper network (e.g., ResNet-34) contains completely the smaller architecture (e.g., ResNet-18) due to the configuration of the same residual blocks. Therefore, it motivates us to think why not directly train a single deep full-net in Fig.~\ref{Fig1}(c) to dynamically switch different-depth sub-nets during the deployment stage.

\par Previous works~\cite{MDN, BYOT, Skipnet} utilize dynamic routines to construct depth-wise dynamic neural networks. ~\cite{SNN, SNN_v2} train a shared network with switchable batch normalization to adjust its width to construct width-level dynamic neural networks. However, above-mentioned notable works can not prove the accuracy of each sub-net. ~\cite{ESD, EKD} introduce knowledge distillation ~\cite{KD} into ensemble network to dynamically switch branches during inference. However, when more sub-nets are required, the total size of ensemble network will become larger and the training process will become more complex.

\begin{figure*}[t]
\begin{center}
\includegraphics[width=0.84\linewidth]{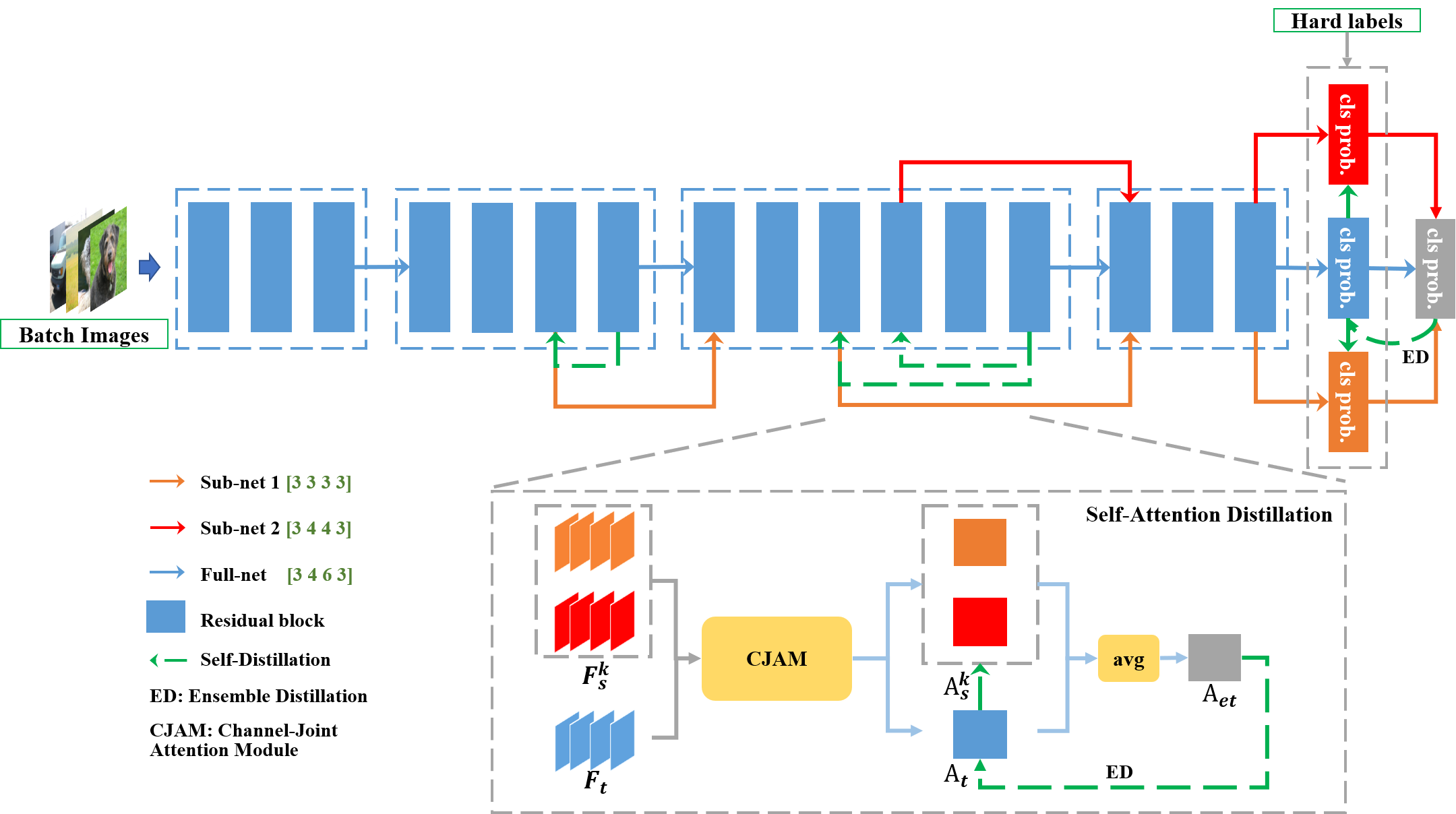}
\end{center}
   \caption{Proposed depth-level dynamic neural network with embedded knowledge distillation training. In this case, full-net is ResNet-50 while sub-nets are ResNet-44 and ResNet-38. [3, 4, 6, 3] means each stage of full-net contain 3, 4, 6 and 3 residual blocks respectively.}
\label{Fig2}
\end{figure*}

\par How to efficiently train one DDNN with multiple high-accuracy sub-nets with simple training strategy? We propose a depth-level dynamic neural network embedded knowledge distillation (EKD-DDNN). As shown in Fig.~\ref{Fig1}(c), we first split the full-net by setting split points in each stage of network. Then, we utilize the predicted class posterior probabilities of full-net as soft labels to guide the learning of different-depth sub-nets. We also introduce the self-attention on the same resolution feature of different-depth to drive the learned semantic feature as consistently as possible between full-net and sub-net. Meanwhile, we still keep the conventional cross-entropy loss term with hard labels for the training of full-net and sub-nets. In addition, we introduce ensemble learning to enhance full-net using ensemble result of full-net and sub-nets as soft-label. Finally, we integrate the above-mentioned strategy and optimize the whole network with EKD mechanism in an end-to-end manner.
\par To verify the effectiveness of proposed method, we conduct extensive experiments on multiple benchmark datasets (e.g., CIFAR-10/100~\cite{CIFAR}, and ImageNet~\cite{ImageNet}) with current state-of-the-art networks (e.g., VGGNet~\cite{VGG}, ResNet~\cite{ResNet}). Compared to the individually training strategies without shared weights, sub-nets in DDNN with EKD training achieve higher performance while preserving the original accuracy of full-net. The main contributions can be summarized as follows.
\begin{itemize}
 \item We propose a Depth-Level Dynamic Neural Network for satisfying the requirement of different resource-constrained devices.
 \item We adopt Embedded-Knowledge-Distillation training mechanism to effectively improve the representative capacity of multiple sub-nets without harming the capacity of full-net.
 \item Sub-nets with EKD improve considerably, which can sometimes surpass individual training networks by more than 1\% on CIFAR and ImageNet.
\end{itemize}

\section{Methodology}
\par In this paper, we focus on constructing depth-level dynamic neural network with EKD training mechanism to excavate the potential representative capacity of sub-nets.
\par \textbf{Depth-Level Dynamic Neural Network.} Depth-Level dynamic neural network denotes that one deep neural network (i.e., full-net) can dynamically switch its depth to yield different-depth sub-nets for different resource-limited devices. Generally, we select a large baseline network as full-net to design the depth-level dynamic neural network. We first separate the large baseline network into several stages according to the resolution of output feature maps. Then, we skip the deep blocks of some stages to create sub-stages. Finally, a sub-net is formed by creating direct connections between sub-stages of each two neighbouring stages. Fig.~\ref{Fig2} uses ResNet-50 as example to show the construction process of DDNN.
\par \textbf{Embedded Knowledge Distillation.} Fig.~\ref{Fig2} shows the detailed training process of our DDNN with the EKD training mechanism. The full-net is used as ``teacher model'' to provide extra supervised information for guiding the better learning of sub-nets. Meanwhile, the full-net also benefits from better sub-nets because of possessing better sub-nets as ``backbone''. Besides, full-net is optimized with ensemble logits and attention maps, which means ``teacher model'' also ``discusses'' from ``student models'' for better learning.
\par As shown in Fig.~\ref{Fig2}, we embed KD on posterior class probabilities and intermediate feature maps. Here, we use the common Kullback-Leibler divergence constraint as distillation loss on posterior class probabilities. The formulation is shown in Eq.~\ref{eq1} and Eq.~\ref{eq2}.

\begin{equation}
  \emph{KL}_{s}^{k} = -\frac{1}{N} \sum_{n=1}^{N} \bm{p}_{t}(\bm{x}_{n}) \text{log} \frac{\bm{p}_{s}^{k}(\bm{x}_{n})}{\bm{p}_{t}(\bm{x}_{n})}
\label{eq1}
\end{equation}

\begin{equation}
  \emph{KL}_{t} = -\frac{1}{N} \sum_{n=1}^{N} \bm{p}_{et}(\bm{x}_{n}) \text{log} \frac{\bm{p}_{t}(\bm{x}_{n})}{\bm{p}_{et}(\bm{x}_{n})}
\label{eq2}
\end{equation}

Where $\bm{p}_{t}(\bm{x}_{n})$ and $\bm{p}_{s}^{k}(\bm{x}_{n})$ respectively denote the posterior class probability of full-net and sub-nets. $\bm{p}_{et}(\bm{x}_{n})$ represents ensemble teacher logits which can be formulated as $\bm{p}_{et}(\bm{x}_{n}) = \frac{1}{K+1}(\sum_{k=1}^{K}{\bm{p}_{s}^{k}(\bm{x}_{n})} + \bm{p}_{t}(\bm{x}_{n}))$. $K$ is the number of sub-nets.
\par To further make the semantic feature in the same stage as consistent as possible between full-net and sub-nets, we introduce the self-attention distillation in intermediate feature maps. We adopt mean-squared-error (MSE) loss in each training step. The formulation is shown in Eq.~\ref{eq3} $\sim$ Eq.~\ref{eq6}.

\begin{equation}
  \bm{A_{t}} = \sum_{c=1}^{C}{\bm{(F_{t})_c}},~~ \bm{A_{s}^k} = \sum_{c=1}^{C}{(\bm{F_{s}^k)_{c}}}
  \label{eq3}
\end{equation}

\begin{equation}
  \bm{A_{et}} = \frac{1}{K+1}(norm(\bm{A_{t})} + \sum_{k=1}^{K}{norm(\bm{A_{s}^k}}))
  \label{eq4}
\end{equation}

\begin{equation}
   \emph{MSE}_s^k = \frac{1}{N} \sum_{n=1}^{N} \emph{MSE} \big{(} A_s^k(\bm{x}_n) \| A_t(\bm{x}_n) \big{)}
\label{eq5}
\end{equation}

\begin{equation}
   \emph{MSE}_t = \frac{1}{N} \sum_{n=1}^{N} \emph{MSE} \big{(} A_t(\bm{x}_n) \| A_{et}(\bm{x}_n) \big{)}
\label{eq6}
\end{equation}

Where $\bm{F_{t}}, \bm{F_{s}^k}$ $\in \mathbb{R}^{C \times H \times W}$ indicate feature maps of full-net and $k^{th}$ sub-nets at same level of model. $\bm{A_{t}}, \bm{A_{s}^k}$ $\in \mathbb{R}^{1 \times H \times W}$ indicate channel-joint attention maps, which will be fused to form ensemble attention teacher ($\bm{A_{et}}$). $norm(\cdot)$ denotes the spatial-wise normalization operation.
\par Our DDNN fulfils the online EKD learning via $\emph{KL}$, $\emph{MSE}$ and conventional cross-entropy loss ($L_{t}$ and $L_{s}^{k}$ for full-net and sub-nets each). The whole objective ($L$) of DDNN with EKD training is defined in Eq.~\ref{eq7}.

\begin{equation}
 \begin{split}
   L &= \underbrace{L_{t} + \frac{1}{K}\sum_{k=1}^{K} L_k}_\text{cross-entropy loss}  + \underbrace{\frac{1}{K+1} \cdot w_k ( KL_{t} + \sum_{k=1}^K{KL_{s}^{k}})}_\text{KL distillation} \\
                 &+ \underbrace{\frac{1}{K+1} \cdot \alpha_k (\emph{MSE}_{t} + \sum_{k=1}^K{\emph{MSE}_{s}^{k}})}_\text{Self-attention distillation}
 \end{split}
\label{eq7}
\end{equation}

Where $w_k$ and $\alpha_k$ are the hyper-parameters to adjust the proportions of $\emph{KL}$ distillation loss and $\emph{MSE}$ self-attention distillation loss respectively.

\begin{table}
  \begin{center}
  \renewcommand{\arraystretch}{1.25}
  \resizebox{1.00\linewidth}{!} {
  \begin{tabular}{c| c c c| c c c}
  \hline
  \multirow{2}{*}{Networks}    &\multicolumn{3}{c|}{CIFAR-10}        &\multicolumn{3}{c}{CIFAR-100}   \\
  \cline{2-7}
                               &Individual    &DDNN           &EKD           &Individual      &DDNN            &EKD   \\
  \cline{1-7}
  ResNet-16 [2 2 2]             &8.87          &8.91           &\textbf{7.99} &33.95           &34.23           &\textbf{32.45} \\
  ResNet-20 [3 3 3]             &7.92          &8.31           &7.94          &32.52           &33.11           &\textbf{32.33} \\
  \cline{1-7}

  ResNet-20 [3 3 3]              &7.92          &8.00           &\textbf{7.34} &32.52           &32.62           &\textbf{30.79} \\
  ResNet-32 [5 5 5]             &7.25          &7.72           &7.28          &30.10           &31.75           &30.24          \\
  \cline{1-7}

  ResNet-32 [5 5 5]              &7.25          &7.15           &\textbf{6.77} &30.10           &29.97           &\textbf{29.17} \\
  ResNet-44 [7 7 7]             &6.49          &6.78           &6.55          &29.23           &29.81           &29.32          \\
  \cline{1-7}

  ResNet-44 [7 7 7]              &6.49          &6.77           &\textbf{6.00} &29.23           &29.16           &\textbf{28.22} \\
  ResNet-56 [9 9 9]             &6.09          &6.24           &6.13          &28.58           &28.76           &\textbf{28.47} \\
  \cline{1-7}

  ResNet-32 [5 5 5]              &7.25          &7.57           &\textbf{6.95} &30.10           &31.40           &\textbf{29.36} \\
  ResNet-44 [7 7 7]              &6.49          &6.62           &\textbf{6.19} &29.23           &29.88           &\textbf{28.97} \\
  ResNet-56 [9 9 9]             &6.09          &6.39           &\textbf{6.04}          &28.58           &29.02           &28.68          \\
  \cline{1-7}

  VGG-11 [1 1 2 2 2]                &7.68          &7.75           &\textbf{7.14} &32.00           &31.73           &\textbf{30.78} \\
  VGG-16 [2 2 3 3 3]                &6.19          &6.42           &6.20          &27.72           &28.42           &27.83          \\
  \cline{1-7}

  VGG-13 [2 2 2 2 2]                 &6.23          &6.26           &6.27          &28.55           &28.65           &\textbf{28.12} \\
  VGG-16 [2 2 3 3 3]                &6.09          &6.08           &6.15          &27.72           &27.64           &27.77          \\
  \cline{1-7}
  \end{tabular}
  }
\end{center}
\caption{Top-1 error rate (\%) on CIFAR-10/100. ``Individual'' means networks training individually. Bold indicates best results.}
\label{Tab1}
\end{table}

\section{Experiments}
\par To evaluate our proposed method, we conduct extensive experiments and exhaustive comparisons on CIFAR-10/100 and ImageNet benchmark datasets.
\par \textbf{CIFAR-10/100 Classification.} Table~\ref{Tab1} shows the detailed comparisons of three cases (Fig.~\ref{Fig1}). The DDNN trained with only the optimization objective of hard labels from full-net and sub-nets (2nd column) has lower performance than independently individually training (1st column) because these weight-sharing blocks are very hard to simultaneously match the multiple different optimization objectives. Experimental results on the CIFAR-10/100 show that sub-nets in DDNN with EKD training (3rd column) possess the average 1$\sim$2\% improvement compared to sub-nets in DDNN with the only hard labels (2nd column). Even compared to the independently trained cases (1st column), sub-nets in DDNN with EKD training still have the 0.5$\sim$1\% lower error rate (Bold in Table~\ref{Tab1}) while preserving the performance of full-net (almost no decline). Especially, EKD mechanism performs stronger on CIFAR-100 than CIFAR-10. It fully displays the effectiveness of the proposed mechanism on more complex task.

\begin{table}
  \begin{center}
  \renewcommand{\arraystretch}{1.25}
  \resizebox{0.45\textwidth}{!} {
  \begin{tabular}{c c c| c c| c c| c}
  \hline
  \multicolumn{3}{c|}{Individual networks} &\multicolumn{2}{c|}{DDNN} &\multicolumn{2}{c|}{EKD-DDNN} &\multirow{2}*{FLOPs} \\
    \cline{1-7}
    Networks   &Params  &Err. &Params      &Err.   &Params    &Err.  \\
    \hline
    ResNet-18  &11.7M  &31.22    &\multirow{2}{*}{21.8M} &30.96   &\multirow{2}{*}{21.8M} &\textbf{29.78} &1.8G  \\
    ResNet-34  &21.8M  &26.73       &         &27.44  &               &26.92          &3.6G  \\
    \hline

    ResNet-26  &16.0M &27.56    &\multirow{2}{*}{25.6M} &28.22    &\multirow{2}{*}{25.6M} &\textbf{26.41} &2.3G \\
    ResNet-50  &25.6M &23.94        &         &25.14   &             &24.12          &3.8G \\
    \hline

    ResNet-32  &17.4M &26.13      &\multirow{2}{*}{25.6M} &27.03   &\multirow{2}{*}{25.6M} &\textbf{25.39} &2.8G \\
    ResNet-50  &25.6M &23.94        &         &24.86    &            &23.99          &3.8G \\
    \hline

    ResNet-41  &18.9M &24.80      &\multirow{2}{*}{25.6M} &25.71    &\multirow{2}{*}{25.6M}  &\textbf{24.07} &3.4G \\
    ResNet-50  &25.6M &23.94        &     &24.69       &         &\textbf{23.83} &3.8G \\
    \hline

    ResNet-38  &21.9M &25.19
                               &\multirow{3}{*}{25.6M}    &26.45
                                             &\multirow{3}{*}{25.6M} &\textbf{24.55} &3.2G \\
    ResNet-44  &23.3M &24.68    &                     &25.32 & &\textbf{24.38} &3.7G  \\
    ResNet-50  &25.6M &23.94    &                     &25.03 & &24.10          &3.8G  \\
    \hline
  \end{tabular}
  }
\end{center}
\caption{Top-1 error rate (\%) on ImageNet (single-crop testing). }
\label{Tab2}
\end{table}

\par \textbf{ImageNet Classification.} Table~\ref{Tab2} gives detailed comparisons of three different training frameworks in terms of network architecture, network parameters, and top-1 error rate. As expected, sub-nets in EKD-DDNNs get improved while almost no decline of full-nets. Compared to individual large model (e.g. ResNet-50), our DDNN can integrate multiple networks into one network without adding extra parameters and FLOPs. During inference, DDNN can dynamically switch the depth to fulfil the deployment of different resource-limited devices.
\par
\begin{figure}[t]
  \centering
  \includegraphics[width=0.75\linewidth]{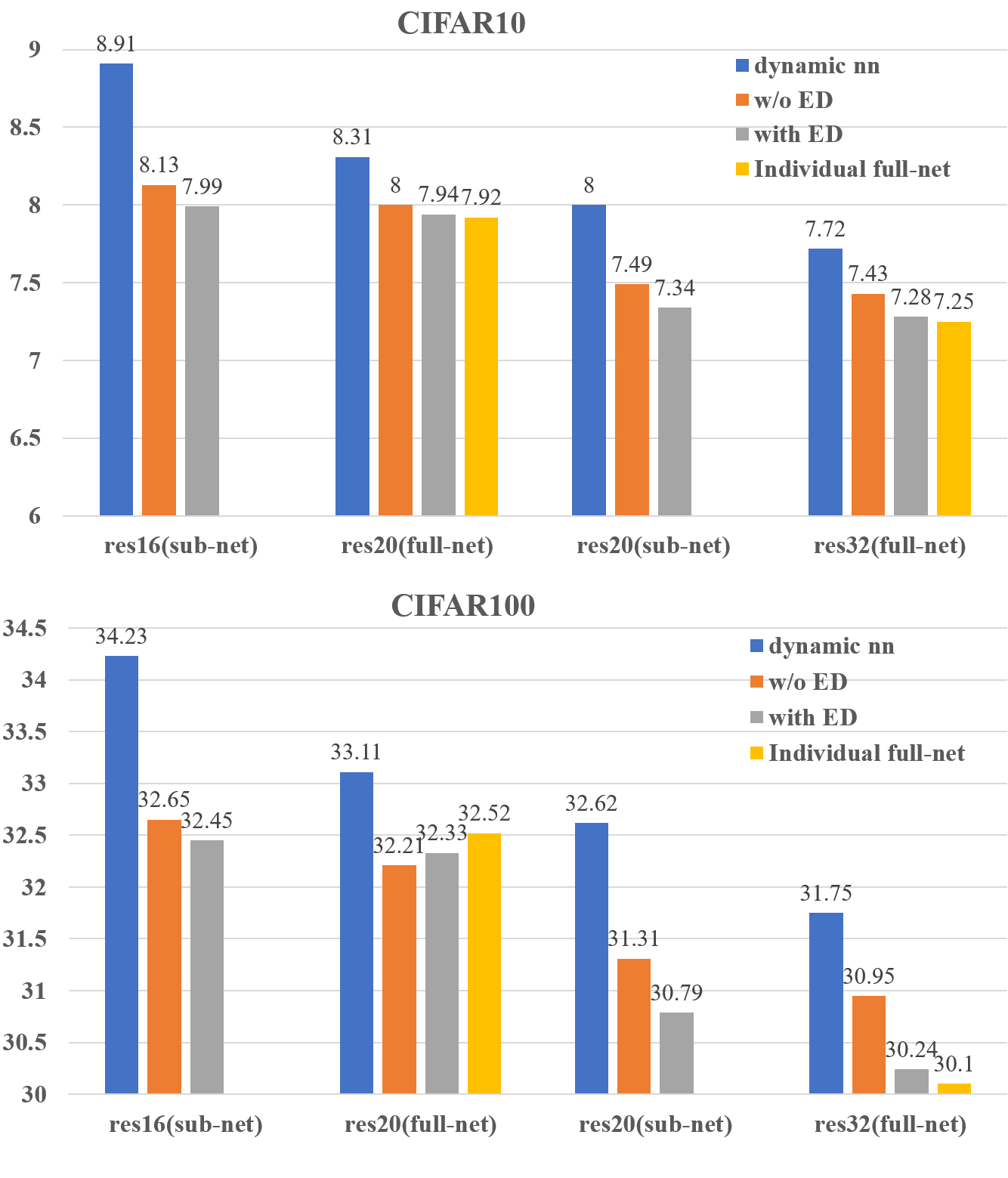}    \\
  \caption{Evaluation on the effectiveness of ensemble distillation.}
  \label{Fig3}
\end{figure}

 \textbf{Effectiveness of Ensemble Distillation.} Using full-net to teach sub-net can enhance the performance of sub-nets. However, it is still hard to guarantee the full-net performance using only a set of weight-sharing blocks. To ease the accuracy decline of full-net, we introduce ensemble distillation on posterior class probability and intermediate feature maps. Fig.~\ref{Fig3} shows that with ensemble distillation, the accuracy of full-net gets maximally preserved.

\section{Conclusion}
In this paper, we propose the DDNN with only one full-net parameters to flexibly switch different-depth sub-nets or full-net according to the demands of different resource-limited devices. To improve the performance of sub-nets and full-net in DDNN, we further design the EKD training mechanism, which contains distillation on posterior class probabilities and self-attention on feature maps to exploit the potential representative capacity of the whole DDNN. Extensive experiments demonstrate that our DDNN with EKD training mechanism achieves competitive performance on multiple benchmark datasets. Compared to student networks trained with previous dynamic paradigm, our sub-nets can harvest better performance with no harm to full-net.

{\small
\bibliographystyle{ieee_fullname}
\bibliography{egbib}

\begin{thebibliography}{10}\itemsep=-1pt

\bibitem{EKD}
Umar Asif, Jianbin Tang, and Stefan Harrer.
\newblock Ensemble knowledge distillation for learning improved and efficient
  networks.
\newblock In {\em ECAI}, volume 325, pages 953--960, 2020.

\bibitem{ImageNet}
Jia Deng, Wei Dong, Richard Socher, Li{-}Jia Li, Kai Li, and Fei{-}Fei Li.
\newblock Imagenet: {A} large-scale hierarchical image database.
\newblock In {\em CVPR}, pages 248--255, 2009.

\bibitem{ResNet}
Kaiming He, Xiangyu Zhang, Shaoqing Ren, and Jian Sun.
\newblock Deep residual learning for image recognition.
\newblock In {\em CVPR}, pages 770--778, 2016.

\bibitem{KD}
Geoffrey~E. Hinton, Oriol Vinyals, and Jeffrey Dean.
\newblock Distilling the knowledge in a neural network.
\newblock {\em CoRR}, abs/1503.02531, 2015.

\bibitem{MDN}
Gao Huang, Danlu Chen, Tianhong Li, Felix Wu, Laurens van~der Maaten, and
  Kilian~Q. Weinberger.
\newblock Multi{-}scale dense networks for resource efficient image
  classification.
\newblock In {\em ICLR}, 2018.

\bibitem{DenseNet}
Gao Huang, Zhuang Liu, Laurens van~der Maaten, and Kilian~Q. Weinberger.
\newblock {Densely Connected Convolutional Networks}.
\newblock In {\em CVPR}, pages 2261--2269, 2017.

\bibitem{CIFAR}
Alex Krizhevsky and Geoffrey~E. Hinton.
\newblock Learning multiple layers of features from tiny images.
\newblock Technical report, 2009.

\bibitem{Faster_RCNN}
Shaoqing Ren, Kaiming He, Ross~B. Girshick, and Jian Sun.
\newblock {Faster R-CNN: Towards Real-Time Object Detection with Region
  Proposal Networks}.
\newblock {\em PAMI}, 39(6):1137--1149, 2017.

\bibitem{FCN}
Evan Shelhamer, Jonathan Long, and Trevor Darrell.
\newblock Fully convolutional networks for semantic segmentation.
\newblock {\em PAMI}, 39(4):640--651, 2017.

\bibitem{VGG}
Karen Simonyan and Andrew Zisserman.
\newblock {Very deep convolutional networks for large-scale image recognition}.
\newblock In {\em ICLR}, 2015.

\bibitem{ESD}
Devesh Walawalkar, Zhiqiang Shen, and Marios Savvides.
\newblock Online ensemble model compression using knowledge distillation.
\newblock In {\em ECCV}, volume 12364, pages 18--35, 2020.

\bibitem{Skipnet}
Xin Wang, Fisher Yu, Zi{-}Yi Dou, Trevor Darrell, and Joseph~E. Gonzalez.
\newblock Skipnet: Learning dynamic routing in convolutional networks.
\newblock In {\em ECCV}, pages 420--436, 2018.

\bibitem{SNN_v2}
Jiahui Yu and Thomas~S. Huang.
\newblock Universally slimmable networks and improved training techniques.
\newblock In {\em ICCV}, pages 1803--1811, 2019.

\bibitem{SNN}
Jiahui Yu, Linjie Yang, Ning Xu, Jianchao Yang, and Thomas~S. Huang.
\newblock Slimmable neural networks.
\newblock In {\em ICLR}, 2019.

\bibitem{BYOT}
Linfeng Zhang, Jiebo Song, Anni Gao, Jingwei Chen, Chenglong Bao, and Kaisheng
  Ma.
\newblock Be your own teacher: Improve the performance of convolutional neural
  networks via self distillation.
\newblock In {\em ICCV}, pages 3712--3721, 2019.

\end{thebibliography}
}

\end{document}